\documentclass[letterpaper, 10 pt, conference]{ieeeconf}  

\IEEEoverridecommandlockouts                              
\usepackage[utf8]{inputenc}
\usepackage[T1]{fontenc}

\usepackage{float}
\usepackage{graphicx}
\usepackage{lipsum}
\usepackage{stfloats}
\usepackage{multicol}
\usepackage{multirow}
\usepackage{amsmath,bm}
\usepackage{etoolbox}

\usepackage{array}

\usepackage{tabulary}
\usepackage{xcolor}

\usepackage{hyperref}
\hypersetup{colorlinks,linkcolor=black}
\hypersetup{colorlinks,urlcolor=black}

\usepackage[caption=false,font=footnotesize]{subfig}

\title{\LARGE \bf
DR-TANet: Dynamic Receptive Temporal Attention Network for Street Scene Change Detection
}

\author{Shuo Chen$^{1}$, Kailun Yang$^{1}$ and Rainer Stiefelhagen$^{1}$
\thanks{This work was supported in part by the Federal Ministry of Labor and Social Affairs (BMAS) through the AccessibleMaps project under Grant 01KM151112, in part by the University of Excellence through the ``KIT Future Fields'' project, and in part by the Hangzhou SurImage Technology Company Ltd.}
\thanks{$^{1}$The authors are with Institute for Anthropomatics and Robotics, Karlsruhe Institute of Technology, 76131 Karlsruhe, Germany (correspondence: kailun.yang@kit.edu).}%
}

\begin{document}

\maketitle
\thispagestyle{empty}
\pagestyle{empty}

\begin{abstract}

Street scene change detection continues to capture researchers' interests in the computer vision community. It aims to identify the changed regions of the paired street-view images captured at different times. The state-of-the-art network based on the encoder-decoder architecture leverages the feature maps at the corresponding level between two channels to gain sufficient information of changes. Still, the efficiency of feature extraction, feature correlation calculation, even the whole network requires further improvement. This paper proposes the temporal attention and explores the impact of the dependency-scope size of temporal attention on the performance of change detection. In addition, based on the Temporal Attention Module (TAM), we introduce a more efficient and light-weight version - Dynamic Receptive Temporal Attention Module (DRTAM) and propose the Concurrent Horizontal and Vertical Attention (CHVA) to improve the accuracy of the network on specific challenging entities. On street scene datasets `GSV', `TSUNAMI' and `VL-CMU-CD', our approach gains excellent performance, establishing new state-of-the-art scores without bells and whistles, while maintaining high efficiency applicable in autonomous vehicles.

\end{abstract}

\section{Introduction}

Scene Change Detection (SCD) aims to compare images captured at different times to identify changes that occur in the image.
Till now, scene change detection has found various application scenarios, such as land cover monitoring~\cite{daudt2018fully}, medical diagnosis~\cite{ronneberger2015u}, urban landscape analysis and autonomous driving~\cite{alcantarilla2016street}\cite{sakurada2017dense}\cite{sakurada2018weakly}.
With the development of learning systems for image classification and semantic segmentation, Convolutional Neural Networks (CNNs) have been widely used in computer vision tasks. The state-of-the-art networks, such as VGG~\cite{simonyan2015very} or ResNet~\cite{he2016deep}, can extract well-learned feature maps to gain excellent performance.
Since the architectural advance of fully convolutional networks~\cite{long2015FCN}, image segmentation can be performed in an end-to-end fashion.
Similar to dense image semantic segmentation, scene change detection also addresses pixel-wise detection.
Because of this, current networks designed for scene change detection are mostly based on CNNs and encoder-decoder-architectures~\cite{daudt2018fully}\cite{alcantarilla2016street}\cite{sakurada2018weakly}\cite{lei2020hierarchical}\cite{prabhakar2020cdnet++}. 

\begin{figure}[h]
    \centering
    \includegraphics[scale=0.15]{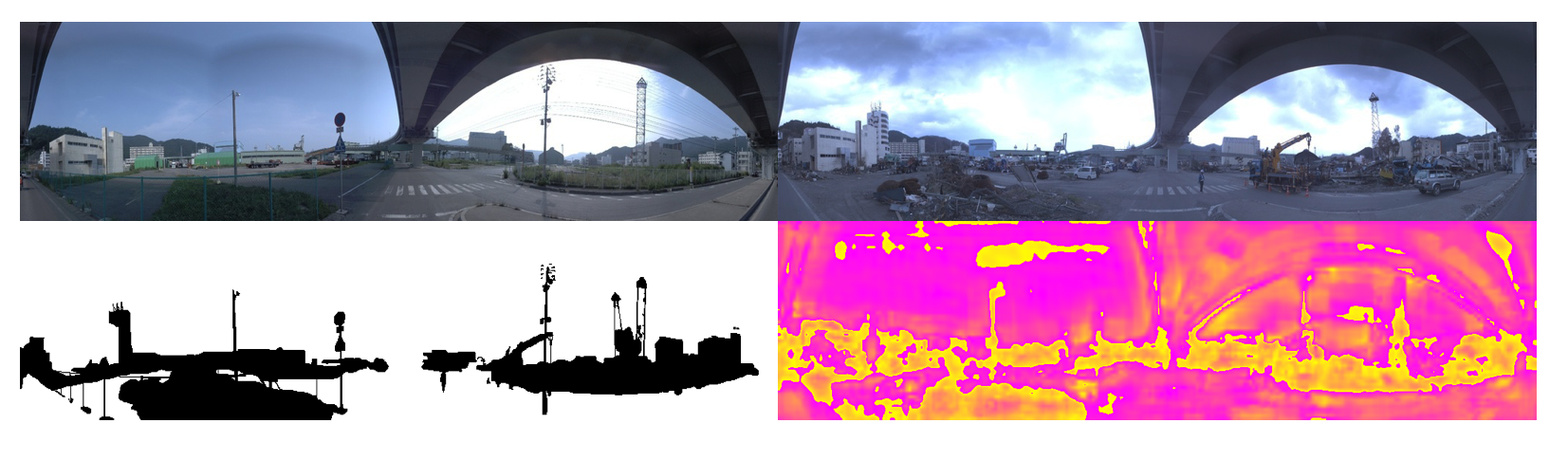}
    \vskip-2ex
    \caption{Illustration of Scene Change Detection. The upper shows paired images at two times. The lower left image is the ground-truth change mask and the lower right image is the visualization of our DR-TANet attention map. The attention map here refers to the similarity map calculated in TAM. The yellow regions in the attention map represent less similar regions.}
    \label{IG}
    \vskip-3ex
\end{figure}

\begin{figure}[t]
    \centering
    \includegraphics[scale=0.23]{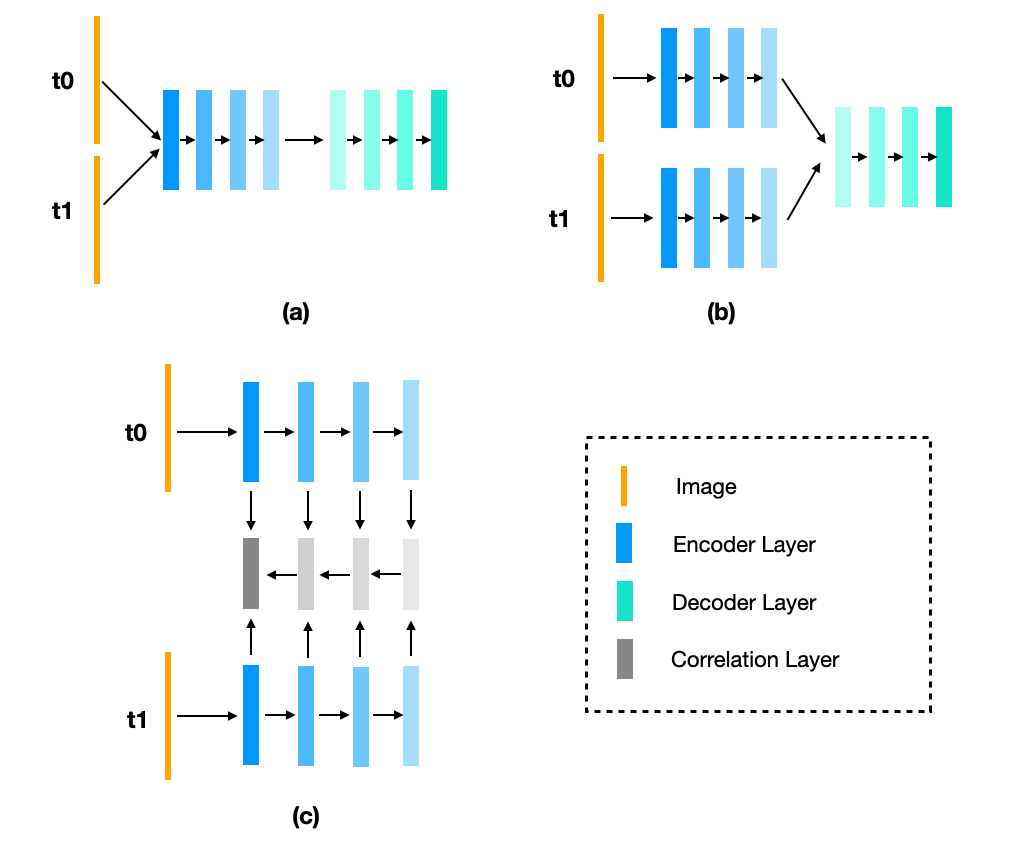}
    \vskip-2ex
    \caption{Fusion mechanisms for Scene Change Detection: (a) Early-Fusion, (b) Late-Fusion, 
    (c) Correlation-Fusion.}
    \label{FT}
    \vskip-4ex
\end{figure}

On account of the specification of scene change detection tasks, as for coping with input images at two temporal points, some frameworks~\cite{daudt2018fully}\cite{alcantarilla2016street}\cite{sakurada2017dense} choose to concatenate the input paired images at the beginning.
In addition, some methods~\cite{sakurada2015change} first extract the feature maps in the two branches and later process the fusion to determine the change areas. 
Further, more research works~\cite{sakurada2018weakly}\cite{lei2020hierarchical}\cite{prabhakar2020cdnet++} have put efforts on approach to fusion of the feature maps extracted from paired images.
These fusion mechanisms are called early-fusion, late-fusion or correlation-fusion, as illustrated in~Fig.~\ref{FT}.

Recently more often, research works~\cite{lei2020hierarchical}\cite{prabhakar2020cdnet++} focus on the relation of feature maps at the same level. The state-of-the-art method is driven by feature correlation~\cite{sakurada2018weakly}\cite{prabhakar2020cdnet++}, which estimates the likelihood of finding the similarity in a fixed neighbourhood, along with paired channel fusion~\cite{lei2020hierarchical}, which combines the same-level feature maps by a cross feature stack to make the channels interweave. However, existing approaches may be incapable of handling objects with various sizes and forms due to their regular and limited sampling ranges. Both quantitative and visual results indicate that the relation of feature maps still has huge research space. From this underground, attention mechanism is introduced in this paper for scene change detection.

In order to achieve better prediction performance, the network for SCD tends to become deeper, such as CSCDNet~\cite{sakurada2018weakly} has the encoder based on the ResNet block, but more deeper, CDNet++~\cite{prabhakar2020cdnet++} inserts five correlation layers, each at the corresponding feature level. However, deeper network will inevitably lead to a lower efficiency. Therefore, the network for SCD requires further work on improving the trade-off between efficiency and performance. Moreover, the estimated change masks in most networks only have the rough outline, yet more detailed change detection should be exploited. 

To address these problems, we propose Temporal Attention Module (TAM), which makes advantages of attention mechanism to find the similarity in a fixed dependency scope.
Besides, we observe that, with the regular square dependency-scope, the strip entities (pedestrian, tree branches, road lamp, etc.) are hard to be detected.
Thereby, we integrate a Concurrent Horizontal and Vertical Attention (CHVA) to perform the refinement.
Finally, we propose the Dynamic Receptive Temporal Attention Module (DRTAM), which is constructed on the basis of TAM, while achieving outstanding performance with the light-weight advantage to make it applicable in intelligent vehicles.

In this paper, the network with the basic Temporal Attention Module (TAM) is called \textbf{TANet} and the network with the Dynamic Receptive Temporal Attention Module (DRTAM), meanwhile refined by the Concurrent Horizontal and Vertical Attention (CHVA), is named \textbf{DR-TANet}.

Our contributions can be summarized as follows:

\begin{itemize}

\item We propose a novel street scene change detection network named TANet and its improved version DR-TANet, which first introduce the attention mechanism into the change detection task. 
On the basis of Temporal Attention Module (TAM), we  explore the performance impact by comparing different dependency-scope sizes and then determine the optimal fixed dependency size.
In addition, because of the increasing receptive field of feature maps in the encoder, we intuitively propose the Dynamic Receptive Temporal Attention Module (DRTAM). DRTAM performs well on public datasets and has less parameters and computation complexity. It improves the efficiency of the whole network and is adopted as the basis of the DR-TANet.

\item We propose the Concurrent Horizontal and Vertical Attention (CHVA), which is integrated into the temporal attention maps to refine the final change detection. CHVA plays an important role on strip entity changes which need to be located accurately for safe driving.

\item Our network outperforms a wide variety of state-of-the-art networks on street scene datasets `GSV' and `TSUNAMI'~\cite{sakurada2015change}. Besides, it performs favorably well on the dataset `VL-CMU-CD'~\cite{alcantarilla2016street}. Our code will be made publicly available at.\footnote{DR-TANet:~\url{https://github.com/Herrccc/DR-TANet}}

\end{itemize}

\section{Related Work}

\subsection{Change Detection}

Detecting regions of change in paired images has always been the interest of computer vision researchers. It can find many applications in diverse scenarios like intelligent vehicle systems. Despite the difference in application scenarios, the research for change detection shares many common techniques and algorithms. Early change detection methods are mainly based on the difference maps and some of them cope with this task with the help of statistical hypothesis test and probability theory~\cite{summarysurvey2005}. More sophisticated methods are exploring the relations of pixel neighbourhoods in the spatial and temporal dimensions. But all of them exploit only information from the raw images~\cite{summarysurvey2005}.

In recent years, with the development of deep learning and machine vision, CNNs are mostly being used in change detection tasks. The state-of-the-art networks are designed with an encoder-decoder architecture~\cite{alcantarilla2016street}\cite{sakurada2017dense}\cite{lei2020hierarchical}. The encoder normally follows state-of-the-art deep networks, such as VGG~\cite{simonyan2015very}, ResNet~\cite{he2016deep} or U-Net~\cite{ronneberger2015u}, with the aim of performing various visual recognition tasks including image classification or semantic segmentation. Throughout the encoder, raw paired images will be downsampled into feature maps of different scales. The extracted features ensure the sufficient usage of the information from the raw images.

The influential and well-performed networks in recent years own the similar architecture. The main difference between these networks lies in the methods of coping with feature map at two temporal channels. In~\cite{sakurada2017dense}, the CDNet based on the U-Net~\cite{ronneberger2015u} adopts the early-fusion mechanism, and the input injected into the encoder, is the concatenated RGB image pair and the estimated dense optical flow. In~\cite{sakurada2018weakly}, the CSCDNet chooses the basic block of ResNet~\cite{he2016deep} as their fundamental feature extraction element. Moreover, two correlation layers are arranged into the model for the estimation of optical flow and stereo matching. CDNet++~\cite{prabhakar2020cdnet++} chooses VGG-19~\cite{simonyan2015very} as the encoder backbone. Furthermore, correlation layers at five feature levels estimate the likelihood of finding a similar feature in a fixed neighborhood.

Despite the advances in SCD, there are still problems to be solved. One network that is both light-weight, highly efficient and able to perfectly detect detailed changes still needs to be developed. Therefore, we introduce DR-TANet, which guarantees excellent performance with fewer parameters and less calculation complexity.

\subsection{Feature Correlation}

As mentioned before, the relation of feature maps at the same level of different temporal channels still needs more exploration.
The state-of-the-art relation calculation method in the task of change detection is correlation calculation. The correlation calculation is processed at a certain feature level~\cite{sakurada2018weakly} or at full feature levels~\cite{prabhakar2020cdnet++}. The correlation maps present the similarity between the certain centered position at t0 and a fixed square search area around the same position at t1.
The correlation mechanism can obtain a relatively good performance, but there is still a large improvement space.

In the field of nature language processing, Transformers~\cite{vaswani2017attention} have been the most popular model in recent years. As the basic module of Transformer, self-attention mechanism helps to learn long-range dependencies~\cite{vaswani2017attention}.
Because of its excellent performance, improved computational complexity and parallelization, self-attention mechanism has been used in visual recognition tasks. In~\cite{huang2019ccnet}, a criss-cross attention module was applied in the semantic segmentation task, which can harvest the contextual information in a very efficient way. In~\cite{hu2019local}, a new image feature extractor was proposed, which is built with self-attention in order to integrate the inferred compositional structure among visual elements in a local area.
As a potential alternative fundamental block in modern vision models, \cite{ramachandran2019stand} shows that self-attention can be a useful replacement for spatial convolutions due to its fewer parameters and floating point operations.

Moreover, in the self-attention mechanism, the covariance between target pixel and each other pixels is calculated with query and key matrices which presents the similarity between the pixel pair. Furthermore, the softmax operation limits the related scope around the target pixel and offers more 
accurate reference of the dependency scope. In comparison with the correlation mechanism in~\cite{sakurada2018weakly}\cite{prabhakar2020cdnet++}, attention mechanism could precisely leverage the surrounding information and indicates the similarity between pixels.
For this reason, we propose the temporal attention to learn the relations of feature maps at two temporal channels to boost the performance.

\begin{figure*}[t]
    \centering
    \includegraphics[scale=0.25]{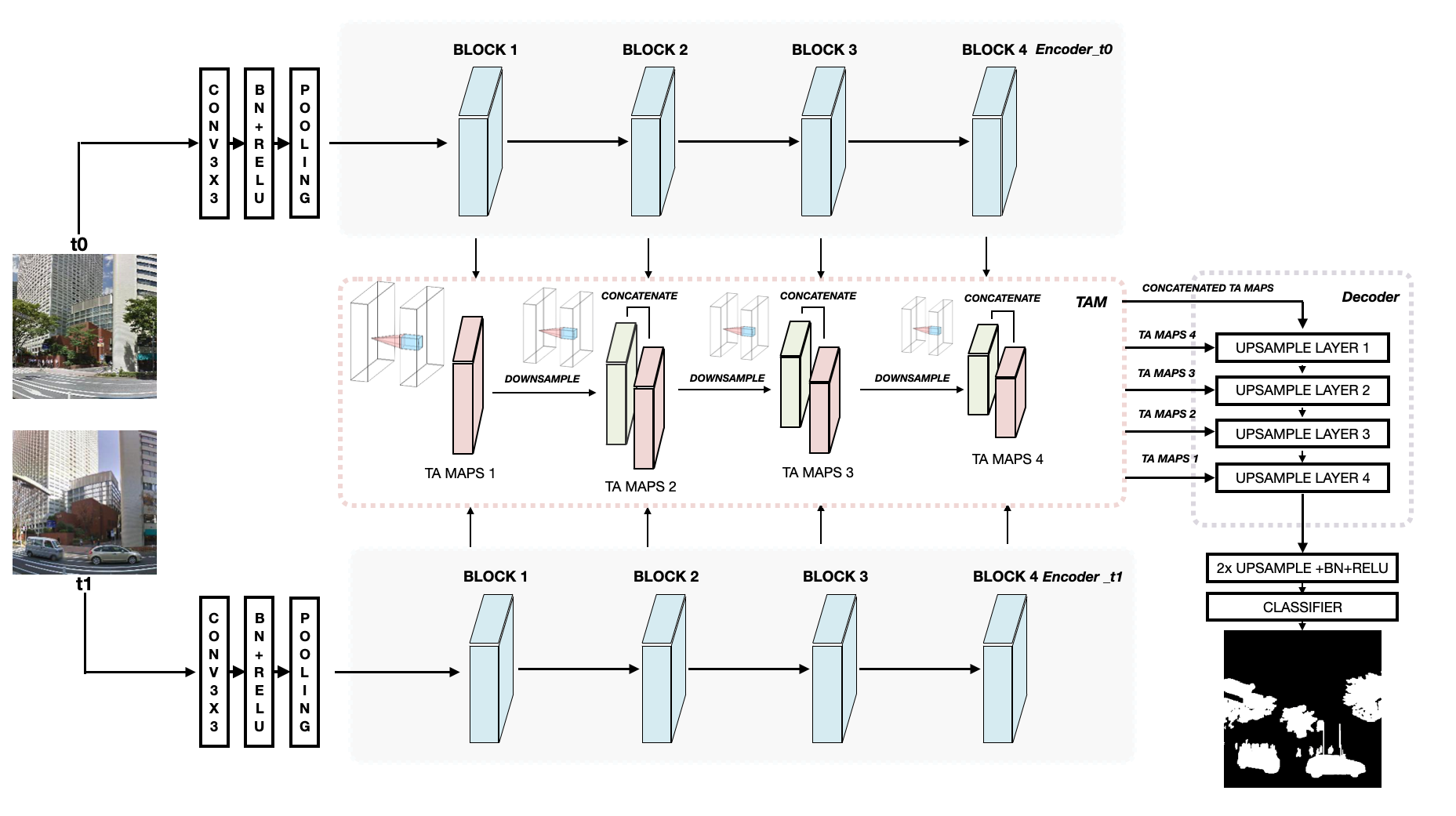}
    \vskip-2ex
    \caption{The overall model architecture of Temporal Attention Network (TANet). The whole model is composed of three main components: encoder (on the left), attention module and decoder (on the right).}
    \label{TAnet}
\vskip-3ex
\end{figure*}

\section{TANet/DR-TANet: Proposed Architecture}

\subsection{Overall Model Architecture}

To achieve efficient change detection, the proposed architecture of TANet has an encoder-decoder architecture, as shown in Fig.~\ref{TAnet}.
The backbone of the encoder is ResNet-18~\cite{he2016deep} to support fast inference.
Due to the paired images at time point t0 and t1, the encoder path has been split into two channels to perform the task of feature extraction.
The extracted feature maps will be injected into the Temporal Attention Module (TAM), with the aim of finding the similarity of feature maps at the corresponding level between two temporal channels.
Finally, the attention maps, which are generated in TAM, will be fed into the decoder.
The decoder is composed of four upsampling layers to perform the upsampling to recover the change mask estimation of the required size.

As the improved version of TANet, DR-TANet abandons TAM and adopts DRTAM for calculating attention maps and CHVA for refinement. DRTAM and CHVA will be detailed in the next subsections.

\subsection{Temporal Attention Module (TAM)}

Inspired by the self-attention mechanism in transformer~\cite{vaswani2017attention} and for vision models~\cite{ramachandran2019stand}, in consideration of the specification of Scene Change Detection (SCD) task, we introduce the Temporal Attention (TA) into the SCD model. 

\begin{figure}[t]
    \centering
    \includegraphics[scale=0.15]{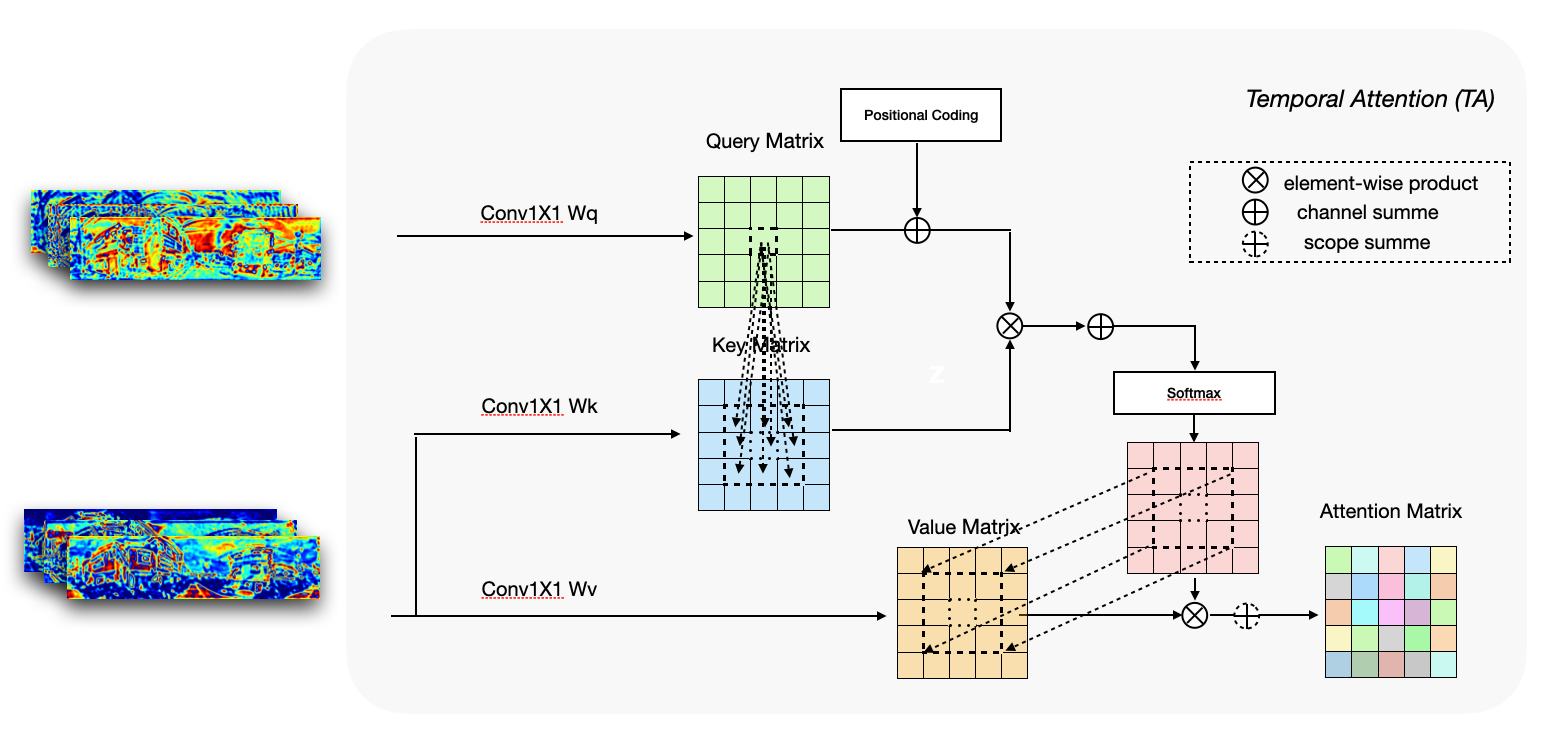}
    \vskip-2ex
    \caption{Schematic of Temporal Attention (TA).}
    \label{TA}
    \vskip-3ex
\end{figure}

During the calculation procedure of Temporal Attention (TA), the feature maps at the t1 channel will be used for generating the query matrix.
The query matrix will be operated with the key matrix to generate the covariance between each pixel from t1 and related pixels in the dependency scope from t0. Covariance matrix after the softmax operation offers accurate similarity between pixels and then operated with value matrix inferred from feature maps at the t0 channel, in order to predict the change regions, as it is shown in Fig.~\ref{TA}.
Vividly, the image at t1 will make an enquiry from the image at t0 about changed regions.

Obviously, if we detect the dependency and similarity between each pixel of the temporal channel t0 and all pixels of the temporal channel t1, we can obtain the full information, however global full-image attention is computationally expensive and not representative.
In this paper, we first apply a fixed dependency-scope size. That means an example pixel (i,j) of feature maps at channel t0 will be regarded to depend only on a fixed scope of pixels of feature maps at channel t1, which center on the same position (i,j).
The fixed dependency-scope size could be (1$\times$1), (3$\times$3), (5$\times$5), etc. The effect of different dependency sizes will be researched in our ablation studies in Section IV.

Moreover, we also conduct the multi-head temporal attention.
That means the feature maps will be divided into N groups in the channel dimension to determine the relation between the channels.

In the transformer model architecture, the positional coding module was designed to inject the relative or absolute positional information.
Similar to that, our TA also introduces the Positional Coding (PC) technique. According to the previous work~\cite{ramachandran2019stand}, the relative positional encoding is the best choice for the vision models.
Thereby, the Temporal Attention (TA), which employs the Positional Coding (PC), models not only the dependency and similarity of two temporal channels, but also the proximity of the local position.
Specifically, the pixel location in the dependency scope will not be treated identically.

Formally, the Temporal Attention is defined as follows:
\begin{align}
q_{ij} &= W_{q}X^{t0}_{ij}, k_{ij} = W_{k}X^{t1}_{ij}, v_{ij} = W_{v}X^{t1}_{ij}\label{eq:TA1}\\
A_{ij} &= \sum_{a,b\in{\boldsymbol{N}(i,j)}}softmax_{ab}\{q_{ij}^{\top}(k_{ij}+e_{ab})\}v_{ab}
\end{align}
where $W_{q}$, $W_{k}$, $W_{v}$ are the weight matrix for generating the query, key and value at pixel position $(i,j)$ and $W_{q}, W_{k}, W_{v}$~$\in$ ${\rm I\!R}^{C_{out}{\times}C_{in}}$. $\boldsymbol{N}(i,j)$ represents the dependency-scope size of t1 feature maps centered at $(i,j)$ pixel position, $e_{ab}$ is the relative positional encoding and query calculates the inner product with the sum of key and positional encoding. The softmax and aggregation operations are processed in the dependency scope $\boldsymbol{N}(i,j)$.

On the basis of the Temporal Attention (TA), Temporal Attention Module (TAM) is constructed. TAM consists of 4 layers and each layer takes feature maps at two temporal channels as input and figures out the attention map based on the TA mechanism.
Due to the different sizes of feature maps, the former calculated attention maps will be downsampled and concatenated with the subsequent attention maps and every attention map will be once more inserted into the decoder through skip connections to prevent information loss during the whole upsampling feature flow.

\begin{figure}[t]
    \centering
    \includegraphics[scale=0.24]{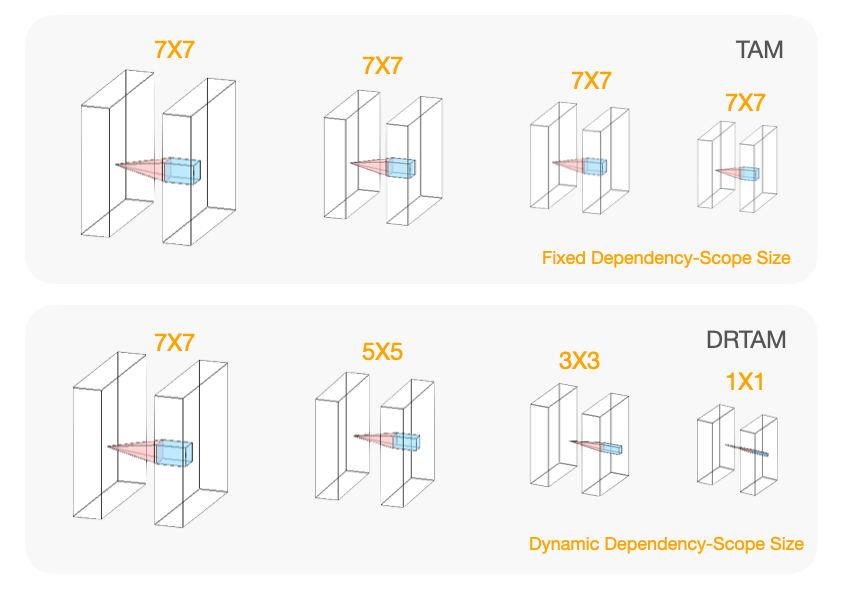}
    \vskip-2ex
    \caption{Top: Temporal Attention Module (TAM) with a fixed dependency-scope size. Bottom: The proposed Dynamic Receptive Temporal Attention Module (DRTAM) with a dynamic dependency-scope size.}
    \vskip-3ex
    \label{DRTAM}
\end{figure}

\subsection{Dynamic Receptive Temporal Attention Module (DRTAM)}

As described before, the TAM is built on a fixed dependency-scope size throughout the whole attention module.
As it is well known that in the vision models, during the downsampling path of the encoder,
feature maps first identify the low-level entities (colour, edges, etc.) and then the high-level entities (texture, shapes, objects, etc.) will gain more focus.
In this regard, intuitively, while
keeping the dependency-scope size unchanged for the constantly downsampled feature maps, the computational capacity is not fully exploited.
With this rationale, in this paper we further propose the Dynamic Receptive Temporal Attention Module (DRTAM), as it is shown in Fig.~\ref{DRTAM}.
When feature maps gather the low-level features, at the initial temporal attention layer, the dependency-scope size will be set at (7$\times$7). Then, at the later temporal attention layers, dependency-scope size will be (5$\times$5), (3$\times$3) and (1$\times$1). In this way, the computational capacity will be highly utilized, while sufficient neighbourhood similarity and dependency will be collected, which are crucial for change detection.

The mathematical expression of DRTAM is as follows:
\begin{align}
&TA_{ab} = softmax_{ab}\{q_{ij}^{\top}(k_{ij}+e_{ab})\}v_{ab}\\
&A^{1}_{ij} = \sum^{a,b\in{\boldsymbol{N_{1}}(i,j)}}_{{\boldsymbol{N_{1}}(i,j)}=7\times7}TA_{ab}, A^{2}_{ij} = \sum^{a,b\in{\boldsymbol{N_{2}}(i,j)}}_{{\boldsymbol{N_{2}}(i,j)}=5\times5}TA_{ab}\\
&A^{3}_{ij} = \sum^{a,b\in{\boldsymbol{N_{3}}(i,j)}}_{{\boldsymbol{N_{3}}(i,j)}=3\times3}TA_{ab},
A^{4}_{ij} = \sum^{a,b\in{\boldsymbol{N_{4}}(i,j)}}_{{\boldsymbol{N_{4}}(i,j)}=1\times1}TA_{ab}
\end{align}
where TA represents the above mentioned temporal attention. $A^{1}$, $A^{2}$, $A^{3}$ and $A^{4}$ represent the TA layer {1,2,3,4} in the DRTAM. $\boldsymbol{N_{x}}(i,j)$ symbolises the dynamically changed dependency-scope size.

\subsection{Refinement: Concurrent Horizontal and Vertical Attention (CHVA)}

\begin{figure}[t]
    \centering
    \includegraphics[scale=0.16]{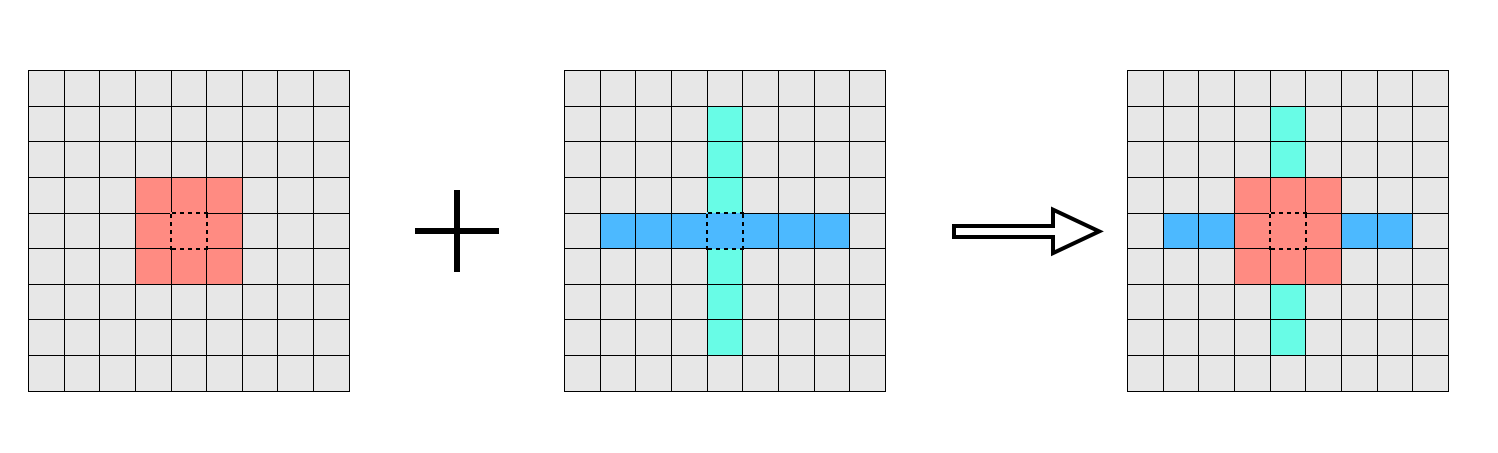}
    \vskip-2ex
    \caption{Illustration of the proposed Concurrent Horizontal and Vertical Attention (CHVA) for change detection refinement.}
    \label{CHVA}
    \vskip-3ex
\end{figure}

Regularly, the convolution in CNNs will define a square around the centered pixel as the associated neighbourhood of centered pixel.
In this paper, we follow this configuration by taking into consideration the shape of dependency scope in TA. 
Yet, the change detection task owns its specifications.
Similar to dense semantic segmentation, it performs pixel-wise detection.
Some objects (tree branches, road lamp) extend horizontally or vertically.
In the research field of semantic segmentation, some researchers have already put efforts on emphasizing the dependencies~\cite{huang2019ccnet}\cite{hou2020strip}\cite{choi2020cars}\cite{ho2019axial}\cite{wang2020axial}\cite{zhang2020semi} along axial directions.
Inspired by these advances and concurrent blocks design~\cite{yang2021context}, we design the Concurrent Horizontal and Vertical Attention (CHVA) to gain more information on both directions for refining the strip entity changes, which is shown in Fig.~\ref{CHVA}.
We adjust the length of horizontal and vertical to (2$\times$K+1), where K stands for the size of the temporal attention dependency scope.

\section{Experiments}

To evaluate the performance of our proposed architecture, we perform experiments on public benchmark datasets - `PCD' (`GSV' and `TSUNAMI') and `VL-CMU-CD'.

\subsection{Street Scene Change Detection Datasets}

Panoramic Change Detection dataset (`\textbf{PCD}' dataset) was first introduced in~\cite{sakurada2015change}.
It was constructed for the street-scene change detection task. As the components of `PCD' dataset, `TSUNAMI' and `GSV' subsets, each of them has one hundred panoramic image pairs and hand-labeled change mask ground truth.
The original size of the image pairs is 224$\times$1024.

\textbf{VL-CMU-CD} dataset~\cite{alcantarilla2016street} consists of 152 perspective image sequences for change detection.
The total amount of image pairs is 1362 and each pair is equipped with manually annotated ground truth.
The original resolution of the images is 1024$\times$768.

\subsection{Experimental Setup}

\subsubsection{\textbf{Dataset Preprocessing}}

During the training process, basically we follow the official configuration of dataset preprocessing. 

For `\textbf{PCD}' dataset, original images are cropped into 224$\times$224.
By sliding 56 pixels in width and data augmentation of plane rotation, each image pair will be expanded into 60 patches with a 224$\times$224 resolution.
In total, 12000 image pairs are generated.
As the input, the image will be resized into 256$\times$256.
We also adopt the 80-20 fold cross-validation setting for training and testing.

For `\textbf{VL-CMU-CD}' dataset, we follow the random training and testing splits in~\cite{alcantarilla2016street}.
933 image pairs (98 sequences) for training and 429 pairs (54 sequences) for testing are resized into a 512$\times$512 resolution. Moreover, the data augmentation technique (plane rotation) will also be applied.

\subsubsection{\textbf{Implementation Details}}

The backbone of our encoder is ResNet-18~\cite{he2016deep}.
During the training procedure, we employ the pre-trained ResNet-18. The upsampling layer is implemented by bilinear interpolation and the classifier layer is convolutional layer with kernel size 1. 
The optimizer Adam~\cite{kingma2014adam} is adopted.
The learning rate is set to 0.001 with $\beta_1$ = 0.9, $\beta_2$ = 0.999. Normally, the batch size is 16. As for the larger dependency-scope size (7$\times$7), due to the limited GPU memory capacity, the batch size will be adjusted to 8. For TAM and DRTAM, the multi-head attention groups number is set to 4. After 100 epochs, the training of TANet converges.
In contrast, the training procedure for DR-TANet will take more epochs to converge until 150 epochs.

\subsubsection{\textbf{Evaluation Metric}}

For the evaluation of the mask map, the evaluation metric F1-score is adopted. The highest value of F1-score is 1 and the lowest is 0.
The higher value means the better precision and recall.
F1-score is calculated upon the fundamental metrics Precision and Recall.
The generalized definition of the metrics are as follows (TP: True Positives; FP: False Positives; FN: False Negatives):
\begin{align}
	Precision &=\frac{TP}{TP+FP} \\
	Recall &=\frac{TP}{TP+FN}\\
	F1-Score &=\frac{2 \times Recall \times Precision}{Recall + Precision}
\end{align}

\begin{table}[t]
\caption{Quantitative comparison with state-of-the-art networks with F1-score. The best result is highlighted in bold.}
\vskip-3ex
\label{tab:result comparison}
\begin{center}
\begin{tabular}{ |ccccc| }
\hline
\multirow{2}{*}{Network} & \multirow{2}{*}{Backbone} &  \multicolumn{3}{c|}{F1-SCORE} \\\cline{3-5}
       &  & TSUNAMI & GSV & AVERAGE\\
\hline
EFnet~\cite{daudt2018fully} & U-Net & 0.659 & 0.565 & 0.612 \\
Siam-Conc~\cite{daudt2018fully} & U-Net& 0.709 & 0.638 & 0.673 \\
Siam-Diff~\cite{daudt2018fully} & U-Net & 0.717 & 0.647 & 0.682 \\
CosimNet~\cite{guo2018learning} & DeepLabV2 & 0.806& 0.692 & 0.749 \\
CDNet~\cite{sakurada2017dense} & U-Net & 0.838 & 0.693 & 0.766 \\
CDNet++~\cite{prabhakar2020cdnet++} & VGG-19 & 0.860 & 0.680 & 0.770 \\
CSCDNet~\cite{sakurada2018weakly} & ResNet-18 & 0.859 & 0.738 & 0.799 \\
\hline
DR-TANet (Ours) & ResNet-18 & \textbf{0.906} & \textbf{0.871} & \textbf{0.889} \\
\hline
\end{tabular}
\end{center}
\vskip-3ex
\end{table}

\subsection{Evaluation on `PCD' Dataset}

Tab.~\ref{tab:result comparison} shows the quantitative comparison with other state-of-the-art change detection networks.
The networks that utilize information from other research fields, such as optical flow and segmentation are not treated as the compared networks.
We focus only on the networks that exploit the feature maps and their relationship at two temporal channels.

\begin{figure}[t]
    \centering
    \includegraphics[scale=0.28]{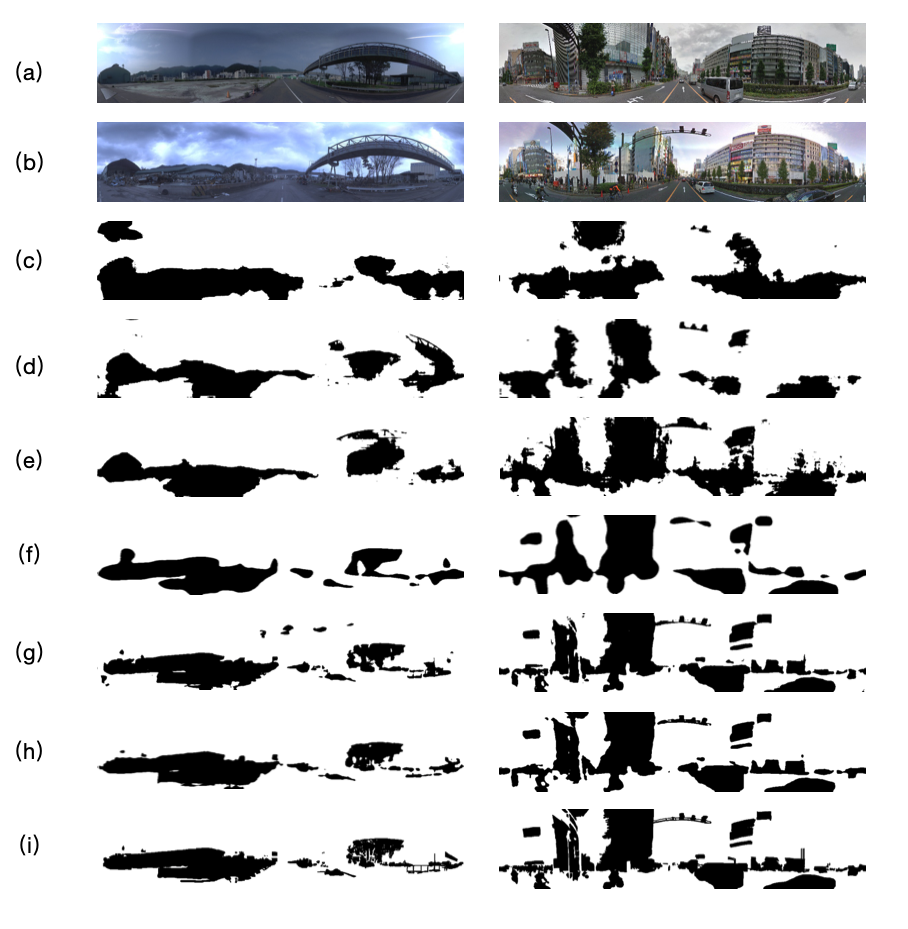}
    \vskip-2ex
    \caption{Visual illustration of change mask estimation of different networks: (a) image at t0, (b) image at t1, (c) EFnet~\cite{daudt2018fully}, (d) Siam-Conc~\cite{daudt2018fully}, (e) Siam-Diff~\cite{daudt2018fully}, (f) CDNet~\cite{sakurada2017dense}, (g) CSCDNet~\cite{sakurada2018weakly}, (h) Our proposed DR-TANet, (i) hand-labeled ground truth.}
    \label{visualres}
    \vskip-3ex
\end{figure}

\begin{figure*}[t]
    \centering
    \includegraphics[scale=0.26]{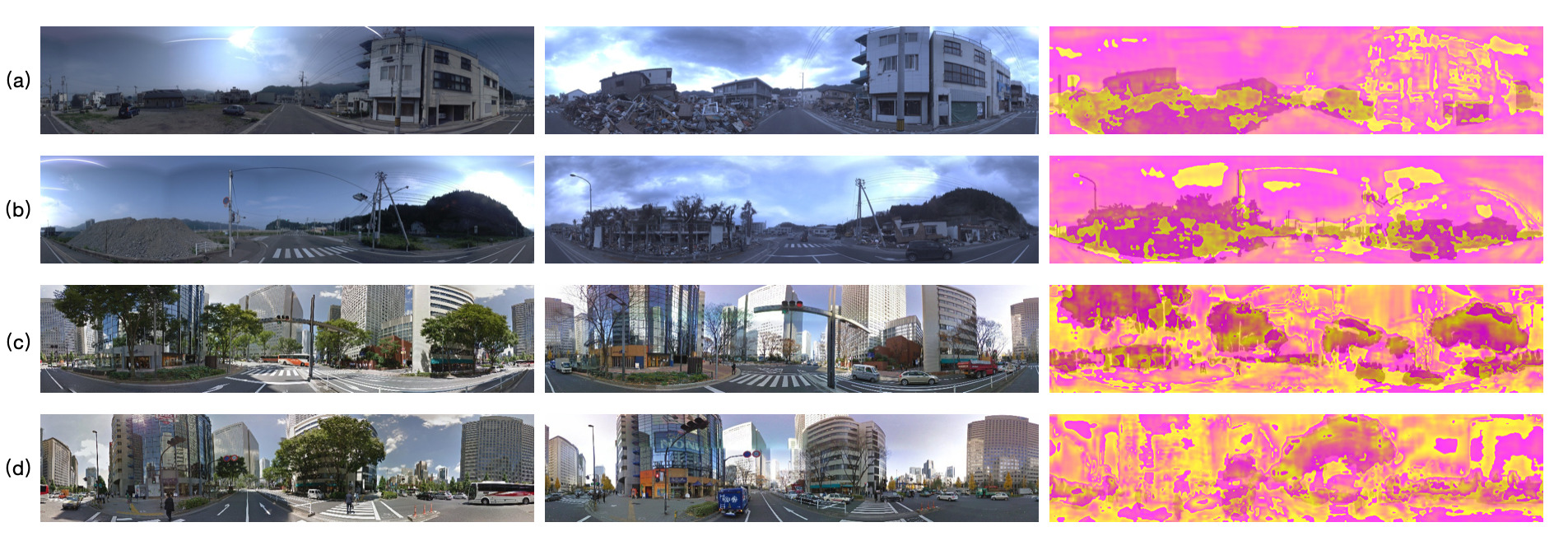}
    \vskip-2ex
    \caption{Illustration of the Temporal Attention (TA) map (overlapped with ground-truth mask). (a) and (b) show the 27-th TA map at TA layer 1; (c) and (d) visualize the 60-th TA map at TA layer 1. At TA layer 1, 64 attention maps are generated. In the 27-th TA map, the highlighted region represents the attended region. On the contrary, in the 60-th TA map, the darker area gets more attention.}
    \label{AM}
\vskip-3ex
\end{figure*}

On the `PCD' dataset, our TANet outperforms all previous SCD networks.
In comparison with CSCDNet~\cite{sakurada2018weakly}, on `TSUNAMI' subdataset, our TANet gains an improvement of 4.7\% in F1-SCORE metric. On `GSV' dataset, our network's performance increases by a large absolute amount of 13.3\%, significantly elevating the state of the art. Visual result comparison is shown in~Fig.~\ref{visualres} and for understanding the mechanism of the TA, TA maps are visualized in~Fig.~\ref{AM}. It can be seen that our DR-TANet is able to capture more detailed and tiny changes, such as tree branches, pedestrians, lamp post, etc.

\subsection{Ablation Studies}

\subsubsection{\textbf{Effect of the TA dependency-scope size in TAM}}

During the calculation procedure of TA maps, the only variable parameter is the dependency-scope size.
For each pixel in feature maps at t0 channel, it determines the range of pixels of feature maps at t1 channel which center at the same pixel position.
The larger the scope size, the wider dependency will be inferred. Undoubtedly, the wider dependency encompasses more detailed local information.

Tab.~\ref{tab:diff.neigh.size} perfectly verifies the initial assumption.
With the (1$\times$1) configuration, the whole network does not perform very well.
The average F1-SCORE on the whole `PCD' dataset stays on a low value 0.781. However, when the dependency size is raised to (3$\times$3), the whole network performance makes a significant improvement, precisely by 7.5\% on `TSUNAMI' subdataset, by 7.7\% on `GSV' subdataset, and by 7.6\% in average. Afterwards, when the scope continues to be raised to (5$\times$5) and (7$\times$7), the performance further slightly increases.

\begin{table}[h]
\caption{Result comparison of different TA configurations. The highest f1-score is in bold and underline.}
\vskip-2ex
\label{tab:diff.neigh.size}
\begin{center}
\begin{tabular}{ |cccc| }
\hline
\multirow{2}{*}{TA dependency-scope size} &  \multicolumn{3}{c|}{F1-SCORE} \\\cline{2-4}
           & TSUNAMI & GSV & AVERAGE\\
\hline
1$\times$1 &  0.807 & 0.755 & 0.781 \\
3$\times$3 & 0.882 & 0.832 & 0.857 \\
5$\times$5 & \underline{\textbf{0.886}} & 0.832 & 0.859 \\
7$\times$7 & 0.885 & \underline{\textbf{0.837}} & \underline{\textbf{0.861}} \\
\hline
\end{tabular}
\end{center}
\vskip-3ex
\end{table}

\subsubsection{\textbf{Effect of Concurrent Horizontal and Vertical Attention (CHVA)}}

\begin{figure*}[h]
    \centering
    \includegraphics[scale=0.26]{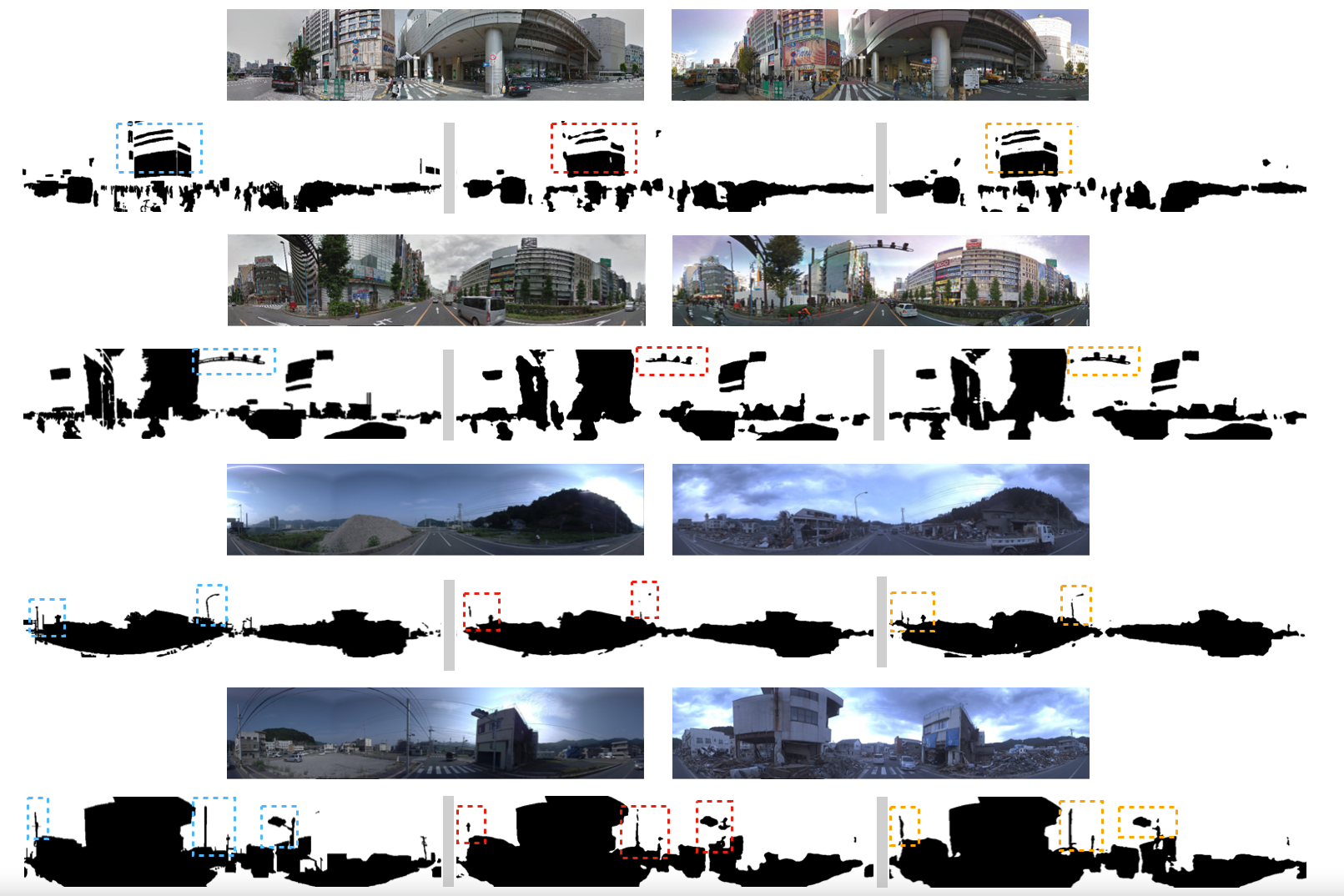}
    \caption{Illustration of refinement with Concurrent Horizontal and Vertical Attention (CHVA). On the top of each group are the input RGB image pairs. The binary change masks are on the bottom. The dark pixels represent the pixels with change occurrence. The left change mask is the hand-labeled ground truth, the middle is the estimation without CHVA and the right is the estimation with CHVA. Dashed rectangles present the hard-detected strip entities.}
    \label{ref of CHVA}
\end{figure*}

As for the Concurrent Horizontal and Vertical Attention (CHVA), the quantitative results in Tab.~\ref{tab:pc/ref} can not show the complete effect of it. The motivation of introducing CHVA is to incorporate wider information along horizontal and vertical directions.
The improvement of estimation on specific image pairs does not have much influence on the whole dataset in terms of the average F1-SCORE.
However, as it is shown in Fig.~\ref{ref of CHVA}, the TA configuration with CHVA achieves clear refinements on the estimation of strip entities, improving the change detection for those challenging areas. 

\begin{table}[t]
\caption{Result comparison with or without CHVA.}
\vskip-2ex
\label{tab:pc/ref}
\begin{center}
\begin{tabular}{ |lccc| }
\hline
\multirow{2}{*}{TA config.} &  \multicolumn{3}{c|}{F1-SCORE} \\\cline{2-4}
           & TSUNAMI & GSV & AVERAGE\\
\hline
3$\times$3  & 0.882 & 0.832 & 0.857 \\
3$\times$3 + CHVA & 0.883 & 0.836 & 0.860\\
\hline
5$\times$5  & 0.883 & 0.831 & 0.857 \\
5$\times$5 + CHVA & 0.884 & 0.836 & 0.860\\
\hline
7$\times$7  & 0.890 & 0.846 & 0.868 \\
7$\times$7 + CHVA & 0.892 & 0.847 & 0.870 \\
\hline
\end{tabular}
\end{center}
\vskip-3ex
\end{table}

\subsubsection{\textbf{Effect of the Dynamic Receptive Temporal Attention Module (DRTAM)}}

As it is shown in Tab.~\ref{tab:res/DRTAM}, the model with Dynamic Receptive Temporal Attention Module (DRTAM) can achieve nearly the same performance as the model with the basic Temporal Attention Module (TAM). Its prominent strength on module complexity therefore makes more significance.

\begin{table}[h]
\caption{Results based on the Dynamic Receptive Temporal Attention Module (DRTAM).}
\vskip-2ex
\label{tab:res/DRTAM}
\begin{center}
\begin{tabular}{ |lccc| }
\hline
\multirow{2}{*}{TA config.} &  \multicolumn{3}{c|}{F1-SCORE} \\\cline{2-4}
           & TSUNAMI & GSV & AVERAGE\\
\hline
DRTAM  & 0.893 & 0.848 & 0.871\\
DRTAM + CHVA & 0.906& 0.871 & 0.889\\
\hline
\end{tabular}
\end{center}
\vskip-3ex
\end{table}

\subsubsection{\textbf{Parameters and MACs Comparison between Different Configurations and Different Networks}}

During the deep learning network construction procedure, the model complexity and the model performance always seem to be a pair of contradictions.
We attempt to make the best trade-off between them.
Our motivation to construct the TANet, is not only for a better change detection performance, but also for an improved effectiveness of the whole architecture. Therefore, it would be necessary to pay attention to the parameters and computational demand. 

First, we compare the number of parameters (Parameters) and the number of Multiply-Accumulation operations (MACs) of the different of different configurations for the TA to give efficiency-oriented recommendations.
Afterwards, we compare Parameters and MACs of different well-performed networks to demonstrate that DR-TANet also has its strength from this perspective.

As it is shown in Tab.~\ref{tab:Params/Flops_config}, as the dependency-scope size of TAM increases, or as the CHVA is integrated into the network, the number of MACs increases accordingly.
As for the specification of attention mechanism, dependency size and CHVA make no influence on the network parameters amount. After we weigh the performance and complexity of diverse configurations, we conclude that the config DRTAM + CHVA earns the recommendation of the most efficient configuration. It achieves nearly the identical results as TAM with a (7$\times$7) dependency size, but it costs even less calculation resources than TAM with a (3$\times$3) scope size.

\begin{table}[t]
\caption{Comparision of Paramaters and MACs (G:giga, M:million) of different configurations of TANet. The recommended configuration is highlighted in bold and underline.}
\vskip-2ex
\label{tab:Params/Flops_config}
\begin{center}
\begin{tabular}{ |lcc| }
\hline
TA Config. & MACs (G) & \#PARAMs (M)\\
\hline
1$\times$1 & 6.685 & 33.390\\
3$\times$3 & 6.719 & 33.390\\
3$\times$3 + CHVA & 7.082 & 33.390\\
5$\times$5 & 6.759 & 33.390\\
5$\times$5 + CHVA & 7.185 & 33.390\\
7$\times$7 & 6.804 & 33.390\\
7$\times$7 + CHVA & 7.293 & 33.390\\
DRTAM & 6.709 & 33.390\\
\underline{\textbf{DRTAM + CHVA}}& 7.055 & 33.390\\
\hline
\end{tabular}
\end{center}
\vskip-5ex
\end{table}

In Tab.~\ref{tab:Params/Flops_net}, we choose three other most well-performed networks to contrast with the proposed DR-TANet on computational complexity. 
It can be seen that DR-TANet is significantly more efficient in terms of Parameters and MACs.

\begin{table}[h]
\caption{Comparision of Paramaters and MACs on different Networks. The least params and macs will be in bold and underline.}
\vskip-2ex
\label{tab:Params/Flops_net}
\begin{center}
\begin{tabular}{ |lcc| }
\hline
Network & MACs (G) & \#PARAMs (M)\\
\hline
CDNet & 39.639 & 39.074 \\
CDNet++ & 187.277 & 43.462 \\
CSCDNet & 42.146 & 94.200 \\
\hline
DR-TANet (Ours) & \underline{\textbf{7.055}} & \underline{\textbf{33.390}} \\

\hline
\end{tabular}
\end{center}
\vskip-3ex
\end{table}

\subsection{Evaluation on `VL-CMU-CD' Dataset}

\begin{table}[t]
\caption{Quantitative result on `VL-CMU-CD' dataset in comparison with state-of-the-art networks with F1-score. The best result is highlighted in bold and underline.}
\vskip-2ex
\label{tab:result comparison CMU}
\begin{center}
\begin{tabular}{ |ccc| }
\hline
Network & Backbone &  F1-SCORE \\
\hline
EFnet~\cite{daudt2018fully} & U-Net & 0.581 \\
Siam-Conc~\cite{daudt2018fully} & U-Net & 0.664 \\
Siam-Diff~\cite{daudt2018fully} & U-Net & 0.625 \\
CosimNet~\cite{guo2018learning} & DeepLabV2 & 0.706 \\
CDNet~\cite{sakurada2017dense} & U-Net & 0.685 \\
CSCDNet~\cite{sakurada2018weakly} & ResNet-18 & 0.710 \\
\hline
DR-TANet (Ours) & ResNet-18 & 0.744\\
DR-TANet (Ours) & \multirow{2}{*}{ResNet-18} & \multirow{2}{*}{\underline{\textbf{0.755}}}\\
(without CHVA) & & \\
\hline
\end{tabular}
\end{center}
\vskip-6ex
\end{table}

In order to verify the general applicability of our DR-TANet, we also conduct experiment on another dataset `VL-CMU-CD'. Our result compared to the best before gains a 4.5$\%$ improvement, which indicates that DR-TANet can also perform  truly well on this dataset. 
As it is shown in Tab.~\ref{tab:result comparison CMU}, the DR-TANet without refining by CHVA performs even better than DR-TANet with CHVA. Part of it is due to the fact that the changes in `VL-CMU-CD' dataset are not so detailed. CHVA, integrating the horizontal and vertical strip information under such circumstances, does not make significant differences. Still, the proposed DR-TANet has exceeded various existing accurate methods by clear margins.

\section{Conclusion}

In this paper, we introduce the attention mechanism to the street scene change detection task. The attention which is here named Temporal Attention (TA) aims to exploit the similarity and dependency of feature maps at two temporal channels. Based on TAM, we propose the Dynamic Receptive Temporal Attention Module (DRTAM), which ensures high performance and meanwhile reduces the calculation complexity. Moreover, we introduce the Concurrent Horizontal and Vertical Attention (CHVA), which performs the targeted refinement for the prediction of strip entity changes. Experiments on `VL-CMU-CD' and `PCD' datasets show that our DR-TANet outperforms other state-of-the-art networks. In consideration of the whole network parameters and computational demand, our network has achieved an excellent balance between accuracy and efficiency, which is ideally suitable for intelligent vehicle systems.

\bibliographystyle{IEEEtran}
\bibliography{bib.bib}

\end{document}